\documentclass[letterpaper, 10 pt, conference]{ieeeconf}  % Comment this line out if you need a4paper

\IEEEoverridecommandlockouts                              % This command is only needed if 
\usepackage[binary-units=true]{siunitx}
\usepackage{times}
\usepackage{amsmath,amssymb}

\usepackage{overpic}
\usepackage{accents}
\usepackage{graphicx}
\usepackage{bm}
\let\labelindent\relax
\usepackage[inline]{enumitem}
\setlist[itemize,1]{leftmargin=0.45cm}
\usepackage{array}
\usepackage{multicol}
\usepackage{multirow}
\usepackage{tabulary}
\usepackage[table, dvipsnames]{xcolor}
\usepackage{booktabs}
\usepackage{subfig}
\usepackage{todonotes}
\usepackage{cite}
\usepackage[hidelinks]{hyperref}
\usepackage{algorithm}% http://ctan.org/pkg/algorithm
\usepackage{algpseudocode}% http://ctan.org/pkg/algorithmicx
\usepackage{breqn}
\usepackage{tcolorbox}
\usepackage{mathtools}
\usepackage{cuted}
\usepackage{stfloats}
\usepackage{balance}
\usepackage{caption}

\usepackage{tcolorbox}
\usepackage{mathtools}
\usepackage{tikz}
\usetikzlibrary{arrows}
\usetikzlibrary{calc}
\usetikzlibrary{arrows,decorations.markings}
\usepackage{tkz-euclide}
\usepackage{nomencl}
\usepackage{pgfplots}
\usetikzlibrary{arrows.meta}
\usepackage{nicefrac}

\renewcommand{\thetable}{\arabic{table}}

\usepackage{nomencl}
\usepackage{xpatch}

\xpatchcmd{\thenomenclature}{%
	\section*{\nomname}% Look for `\section*... etc.
}{% Replace it by 'nothing'
}{\typeout{Success}}{\typeout{Failure}}
\usetikzlibrary{patterns}
\newcommand{\stripedbox}[1]{%
	\tikz[baseline=(X.base)]{
		\node[
		inner sep=1pt,
		pattern=crosshatch,
		pattern color=gray!20,
		draw=gray!30
		] (X) {#1};
	}
}

\makenomenclature

\renewcommand{\thetable}{\arabic{table}}
\allowdisplaybreaks

\DeclareMathOperator{\diag}{diag}

\definecolor{light-gray}{gray}{0.95}
\definecolor{dark-gray}{gray}{0.5}
\definecolor{mygray}{gray}{0.75}
\newcommand{\BIN}{\begin{bmatrix}}
\newcommand{\BOUT}{\end{bmatrix}}

\newcommand{\hz}{\hat{z}}

\newcommand{\bb}{\breve{b}}

\newcommand{\ta}{\widetilde{a}}
\newcommand{\tA}{\widetilde{A}}

\newcommand{\mA}{{\mathcal{A}}}

\newcommand{\mI}{{\mathcal{I}}}

\newcommand{\bC}{{\mathbb{C}}}
\newcommand{\bD}{{\mathbb{D}}}

\newtheorem{prop}{Proposition}

\pgfdeclarelayer{bg1}   
\pgfdeclarelayer{bg}  
\pgfsetlayers{bg1,bg,main}  

\definecolor{orange}{rgb}{0.99,0.69,0.07}
\definecolor{lightgray}{gray}{0.85}
\definecolor{light-gray}{gray}{0.95}
\definecolor{dark-gray}{gray}{0.5}

\tikzset{cross/.style={cross out, draw=black, minimum size=2*(#1-\pgflinewidth), inner sep=0pt, outer sep=0pt},
cross/.default={1pt}}

 \makeatletter
 \newcommand\fs@spaceruled{\def\@fs@cfont{\bfseries}\let\@fs@capt\floatc@ruled
   \def\@fs@pre{\vspace{5pt}\hrule height.8pt depth0pt \kern2pt}%
   \def\@fs@post{\kern2pt\hrule\relax}%
   \def\@fs@mid{\kern2pt\hrule\kern2pt}%
   \let\@fs@iftopcapt\iftrue}
 \makeatother

\makeatletter
\newcommand{\thickhline}{%
	\noalign {\ifnum 0=`}\fi \hrule height 1pt
	\futurelet \reserved@a \@xhline
}
\newcolumntype{"}{@{\hskip\tabcolsep\vrule width 1pt\hskip\tabcolsep}}
\makeatother

\title{\LARGE \bf Large-Scale Hierarchical Decision-Making in Robot Control via ADMM}

\author{
	Kai Pfeiffer$^{1}$\hspace{-1pt},
	Jia Wang$^{1}$\hspace{-1pt},
	Gordon Owusu Boateng$^{1}$\hspace{-1pt},
	Yuquan Wang$^{2}$\hspace{-1pt},
	Vincent Bonnet$^{3,4}$\hspace{-1pt},
 Abderrahmane Kheddar$^{5}$%
	\thanks{$^{1}$AIAT,
		Xi'an Jiaotong-Liverpool University, China.}%
	\thanks{$^{2}$X Square Robot, China.}%
	\thanks{$^{3}$IPAL, CNRS, Singapore}
	\thanks{$^{4}$LAAS-CNRS, University of Toulouse, France.}%
	\thanks{$^{5}$LIRMM-CNRS, University of Montpellier, France.}%
}

\begin{document}
	
	\maketitle
	\thispagestyle{empty}
	\pagestyle{empty}
	
\begin{abstract}%		
Sparse hierarchical nonlinear programming (\mbox{SH-NLP}) based on the $\ell_0$-norm provides a powerful framework for integrated, efficient hierarchical decision-making in robot kinematic control, for example in the presence of a large set of possible end-effector candidate locations. While Sequential Sparse Hierarchical Quadratic Programming (S-SHQP) with interior-point method based SHQP sub-problem solvers exploit the hierarchical, sparse structure of such problems, they are less amenable to effective warm-starting between subsequent, coherent control steps. 
In this work, we formulate an Alternating Direction Method of Multipliers (ADMM) to efficiently solve related SHQP sub-problems. We introduce efficient substitutions for auxiliary variables representing $\ell_0$-norm surrogates. The solver efficiently handles separable decision functions by evaluating linear algebra expressions in the appropriate order. 
Results in large-scale kinematic and dynamic control show reduced computation times of up to 23.5\% compared with a structure-exploiting interior-point method in low- to moderate-accuracy regimes.
Furthermore, hierarchical decision-making is extended to whole-body dynamic control for humanoid walking, where the need for high-accuracy solutions is highlighted. 
\end{abstract}

\section{Introduction}

Hierarchical decision-making in robot control frequently requires selecting among a large number of candidate actions or locations while simultaneously satisfying whole-body kinematic and physical constraints of different priorities. Examples include selecting footholds~\cite{Song2021}, or activating only a subset of joints in highly redundant robotic systems~\cite{Dinev2021}, all while respecting higher or lower priority constraints~\cite{Yu2026}. Sparse hierarchical nonlinear programming (SH-NLP)~\cite{pfeiffer2026} provides an attractive formulation for such problems, as discrete decisions can be embedded directly into continuous programs without explicitly introducing combinatorial decision variables.

SH-NLPs can be addressed by iteratively solving sparse hierarchical quadratic programs (SHQP), for example by the interior-point method (IPM)~\cite{pfeiffer2026}. By exploiting the sparse hierarchical structure, a high degree of efficiency with linear dependency in the number of sparse candidate constraints can be achieved, unlike standard QP solvers with cubic dependency.

However, exploitable structure exists also between individual SHQP subproblems. Robot control typically consists of high-frequency updates with gradual changes between subsequent instances, for example similar sets of active and inactive inequality constraints. The IPM is less amenable to exploiting this additional sequential coherence through warm-starting.

	\begin{figure}[tp!]
	\includegraphics[width= \columnwidth]{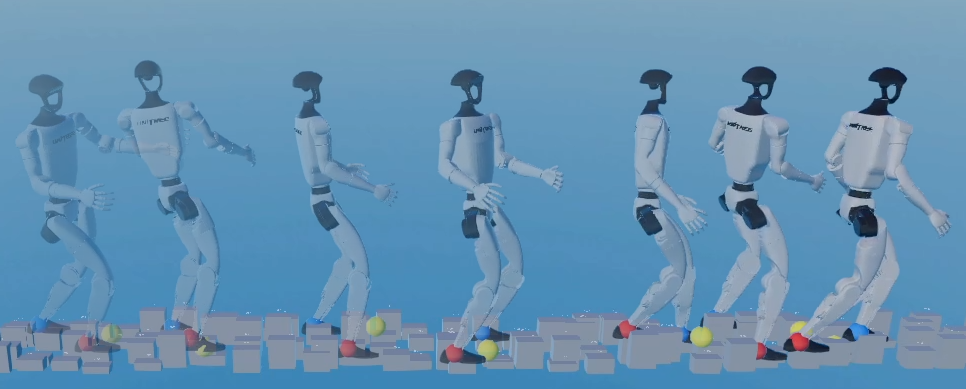}
	\centering
	\caption{Unitree G1 whole-body control with integrated hierarchical decision-making for walking over 200 stepping stones (current decision: yellow sphere; Sec.~\ref{sec:g1walk}).}
	\label{fig:g1walk}
\end{figure}

The Alternating Direction Method of Multipliers (ADMM)~\cite{boyd2011} is a suitable candidate for such parametric programs~\cite{Mahajan2026}, as its primal and dual can be effectively initialized from previous optimization instances. However, specialized solver formulations for SHQPs do not exist.
This work aims to close this gap with the following main contributions:
\begin{itemize}
	\item We derive a structure-exploiting ADMM solver for SHQPs which eliminates sparsity-related auxiliary variables for linear dependence on the number of decision candidates, while enabling warm-starting, unlike existing IPM based methods.
		\item For separable decision functions, we exploit their structure and implement computationally favorable linear algebra evaluations, which improves efficiency for large candidate sets.

  \item We evaluate the proposed solver on large-scale sparse programs and robot-control problems, isolating the effects of sparse structure, warm-starting, and separable decision-function evaluation. The results demonstrate reduced computation times over the structure-exploiting IPM~\cite{pfeiffer2026} in low- to moderate-accuracy regimes.
	
	\item We investigate the extension of hierarchical decision-making from kinematic to dynamic robot control.
	The experiments demonstrate efficient drone control, while a humanoid walking experiment exposes the need for high-accuracy solutions, which is less suitable for first-order methods like ADMM.
\end{itemize}

The remainder of this paper is organized as follows.
Sec.~\ref{sec:relwork} reviews related work on optimization methods for hierarchical decision-making in robot control.
Sec.~\ref{sec:sshqp} recalls the SHQP formulation arising within S-SHQP.
Sec.~\ref{sec:nadm} derives the proposed efficient ADMM solver, and compares it with other SHQP solvers.
Sec.~\ref{sec:admmcomponents} discusses equilibration and warm-starting, while
Sec.~\ref{sec:sepdm} outlines how structure in separable decision functions can be exploited.
Sec.~\ref{sec:eval} evaluates the computational performance and scalability of the proposed solver on numerical and robot control problems.

\section{Related Work}
\label{sec:relwork}

\subsection{Hierarchical Decision-Making in Robotics}

Mixed-integer formulations provide a powerful tool for globally optimal decision-making in robot planning~\cite{Shirai2025}, but their combinatorial complexity can limit their applicability to large candidate sets and real-time control. 
Instead, sparsity-promoting continuous $\ell_1$-norm formulations have been efficiently applied to parsimonious kinematic control~\cite{Goncalves2015} and footstep selection~\cite{Song2021}. However, these methods either rely on specialized algorithms like the simplex method or additional brute-force layers to achieve accurate sparsity.

The approach in~\cite{Candes2007} enhances the sparsity-promoting qualities of the $\ell_1$-norm by sequentially reweighting the linear objective according to a logarithm-based  $\ell_0$-norm surrogate. The $\ell_0$-norm directly measures cardinality and therefore provides a natural representation of sparse decision-making, but needs to be smoothly approximated for continuous optimization.
While~\cite{Candes2007} is limited to linear unconstrained problems, the approach has been extended to hierarchical nonlinear decision-making by incorporation within a SQP-type framework (Sequential Quadratic Programming~\cite{gill2005}; from hereon referred to as Sequential Sparse Hierarchical Quadratic Programming (S-SHQP)~\cite{pfeiffer2026}).
The corresponding dedicated IPM exploits the sparse hierarchical structure of the SHQP sub-problems, resulting in linear computational complexity with the number of candidate constraints. However, warm-starting is not exploited, and the solver has only been applied to purely kinematic control scenarios.

This work addresses this gap and proposes an ADMM solver which can be effectively warm-started for the efficient repeated solution of subsequent SHQP subproblems. Furthermore, hierarchical decision-making is integrated within a humanoid whole-body controller under consideration of instantaneous robot dynamics.

\subsection{Parametric Quadratic Programming}

Repeatedly solving closely related parametric optimization problems is fundamental to real-time robot control. 
Active-set methods like qpOASES~\cite{Ferreau2014} are well suited to exploit this similarity to accelerate subsequent optimizations, because the active set of a preceding QP often provides a good estimate for that of the subsequent problem in high-frequency control. However, active-set methods can suffer from issues like repeated constraint activation and deactivation (cycling) when constraints are linearly dependent, leading to long computation times.

Interior-point methods are numerically robust alternatives to the active-set method, for example by integrating proximal regularization which enables stable convergence even in the case of linearly dependent constraints~\cite{piqp}.
An extension to
SHQPs, which are a member of the broad class of QPs, has been proposed in~\cite{pfeiffer2026}. Linear dependence on the number of decision candidates as in the unconstrained case~\cite{Koh2007} is maintained by efficient exploitation of the underlying sparse hierarchical structure introduced by $\ell_0$-norm surrogates.

Warm-starting interior-point methods is less direct than for active-set methods. 
Some techniques modify previous solutions to meet central feasibility conditions before restarting the method~\cite{Gondzio2008}, which limits the direct use of a preceding solution.

\subsection{Alternating Direction Method of Multipliers (ADMM)}

First-order methods such as ADMM update primal and dual variables through gradient and projection operations with factorization reuse at the cost of reduced accuracy, rather than repeatedly forming and factorizing KKT systems as required by second-order methods such as the IPM. Their computational efficiency can be further improved through effective warm-starting and by caching factorizations and other quantities which remain invariant across inner iterations~\cite{osqp} and consecutive control steps~\cite{Mahajan2026}.
ADMM has been successfully deployed as an SQP subproblem solver for complex whole-body trajectory optimization, while efficiently exploiting temporal sparsity of optimal control problems~\cite{Jordana2025}.

In this work, we adapt the ADMM to the specific sparse and hierarchical structure of SHQPs in order to maintain linear dependency of the solver in the number of candidates in hierarchical decision-making.

\section{Background}
\label{sec:sshqp}

\subsection{Sparse Hierarchical Nonlinear Programming for Hierarchical Decision-Making}

\begin{figure}[htp!]
	\includegraphics[width=1 \columnwidth]{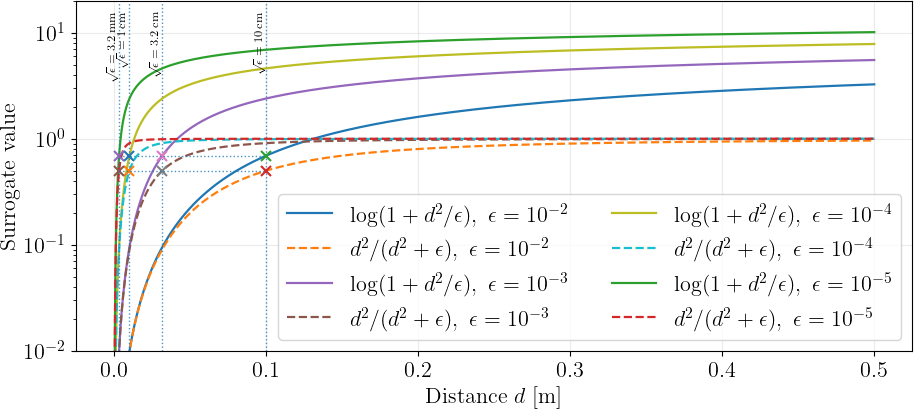}
	\centering
	\caption{Comparison between continuous $\log$~\cite{Candes2007} and fraction function surrogates~\cite{Li2020} of the $\ell_0$-norm on squared decision functions $d^2$ (see~\eqref{eq:dmfct}).}
	\label{fig:l0surrogate}
\end{figure}
This work formulates a continuous optimization problem for hierarchical  decision-making with a more aggressive sparsity promoting approximation of the $\ell_0$-norm with strong gradient decay~\cite{Li2020} (see Fig.~\ref{fig:l0surrogate}). This can be advantageous in the presence of a large number of candidates, when perpetual influence as seen for the logarithmic penalty $\textstyle\sum \log(\tau_d^+\xi)$\cite{Candes2007,pfeiffer2026} is undesired. For completeness, and unlike the work in~\cite{pfeiffer2026}, we also consider $\ell_2$-norm objectives:

\begin{align}
	\min_{\chi,(\nu_d)_{d\in\bD_l},(\tau_d)_{d\in\bD_l},(\nu_i)_{i\in\bC_l}}& 
	\sum_{d\in\bD_l}\nolimits \frac{\tau_d}{\tau_d + \xi} + \left\Vert (\nu_i) \right\Vert_2^2 	\label{eq:shnlp}\tag{S-HNLP}\\
	\text{s.t.}\quad
	-\tau_{d^-}\leq \nu_{d^{\pm}}\leq \tau_{d^+}&,
	\quad d^{\pm}\in\bD_{l}^{\pm},
	\quad l=1,\dots,p\nonumber\\
	f_{d/i/j/k}(\chi) \geqq \nu_{d/i/k}^{(*)}, \quad& d/i/j/k\in\bD_l/\bC_l/\mathcal{I}_{\cup l-1}/\mathcal{A}_{\cup l-1}\nonumber
\end{align}
$\chi\in\mathbb{R}^n$ is the optimization variable, for example representing the robot joint state. The functions $f_{d/i/j/k}(\chi)$ denote nonlinear robot constraints associated with the index sets $d\in\bD_l$, $i\in\bC_l$, $j\in\mathcal{I}_{\cup l-1}$, or $k\in\mathcal{A}_{\cup l-1}$. Each scalar entry $d$ in $\bD_l$ represents a decision candidate, for example a squared expression of kinematic translational reaching $g(\chi)$ with a constant desired value $g_d$ (see Sec.~\ref{sec:sepdm} on how this separable structure can be exploited computationally)
\begin{align}
	f_d(\chi) 
	&= 
	\frac{1}{2}\left\Vert
	g(\chi) - g_d 
	\right\Vert_2^2, \qquad d\in\mathbb{D}_l
	\label{eq:dmfct}
\end{align}
The set $\bC_l$ contains least-squares tasks, while inactive or active tasks from higher priority levels are represented in $\mI_{\cup l-1}/\mA_{\cup l-1}
\coloneqq
\mI_1/\mA_1 \cup \cdots \cup \mI_{l-1}/\mA_{l-1}
$, respectively. These sets may include both equality and inequality constraints, collectively denoted by the relation $\geqq$.

The constraint sets $\bD_l$ and $\bC_l$ are associated with a slack variable $\nu_{d/i}$, which permits relaxation when exact satisfaction is not possible, for example when a task specifies an unreachable target. On each priority level $l$, the slack is optimized until the optimal one $\nu_{d/i}^*$ is found, in which case the associated constraints enter the active set of the next level $\mathcal{A}_{\cup l}$ with fixed optimal slack $\nu_{k}^*$. 
The auxiliary variable $\tau_d$, together with the appropriate bound constraint set $\bD_l^{\pm}$ on $\nu_d$, provides a smooth representation of the element-wise magnitude $|\nu_d|$. For compactness, parentheses denote the stacked vector of quantities indexed by a given set, for example
$(\nu_d)_{d\in\bD_l}
=
\begin{bmatrix}
	\nu_{d_1} & \nu_{d_2} & \cdots
\end{bmatrix}^{T}
\in\mathbb{R}^{|\bD_l|}$:

\textit{In the following, the explicit set memberships $d\in\bD_l$, $i\in\bC_l$, $j\in\mI_{\cup l-1}$, and $k\in\mA_{\cup l-1}$ are omitted whenever they are clear from the context, for example $(f_d) \coloneqq (f_d)_{d\in\bD_l}$}.

\subsection{Sequential Sparse Hierarchical Quadratic Programming}

\ref{eq:shnlp} is solved by iteratively solving~\ref{eq:shqp} subproblems~\cite{pfeiffer2026} for the approximate step variables $x^{(\iota)}$, $v^{(\iota)}$, and $t^{(\iota)}$. The step $x^{(\iota)}$ is subject to a trust region of radius $\phi$  for globalization. In this work focusing on instantaneous robot control, the trust region is kept constant while a filter framework as in~\cite{fletcher2002b,pfeiffer2024} is omitted. SHQP is a constrained version of the reweighted majorization-minimization \(\ell_1\)-optimization problem of \cite{Candes2007}: 

\begin{align}
	\min_{{z_l},(v_d),({t_d}),({t_i})} \quad& \textstyle\sum_{d\in\bD_l}\omega_d{t}_d + \left\Vert (v_i)\right\Vert_2^2\quad l=1,\dots,p\nonumber\\[-2pt]
	&
		{+ 0.5}(x_{l-1}^{*} + N_{l-1}z_l)^T H_l (x_{l-1}^{*} + N_{l-1}{z}_l)
		\nonumber\\
	\text{s.t.}\qquad&-{t}_{d^-} \leq {v}_{d^{\pm}}    \leq{t}_{d^+} ,\qquad d^{\pm}\in\bD_{l}^{\pm} \nonumber \\
	&\ta_d^T{{z}_l} - \bb_d \geqq {v}_d
	\quad d\in\bD_l,\nonumber\\
	&\ta_{i}^T{{z}_l} - \bb_{i} \geqq {v}_{i}
\quad i\in\bC_l\nonumber\\
	&\ta_{j}^T{{z}_l} - \bb_{j}\geq 0,\quad j\in\mathcal{I}_{\cup l-1}\nonumber\\
			&	\vert x_l\vert_{} \leq \phi\label{eq:shqp}\tag{SHQP}
\end{align}
The weights are defined as follows, with a small constant $\xi\approx1\cdot 10^{-4}$ which improves numerical robustness:
\begin{equation}
	\label{eq:omega}
	\omega_d =  \xi/({{\tau}_d + \xi})^2 > 0
\end{equation}
	The positive-definite hierarchical Lagrangian Hessian ${H}_l\succ 0$~\cite{pfeiffer2023} captures second-order terms of the constraints and objectives, while $a_d=\nabla f_d(\chi^{(\iota)})$ and $b_d \coloneqq f_d(\chi^{(\iota)})$ represent linearizations of~\mbox{\ref{eq:shnlp}} at the current iterate $\iota$.
Following the approaches in~\cite{dang2017}, \cite{pfeiffer2026}, we introduced the change of variables:
$
	{x}_l = {x}_{l-1}^* + N_{l-1}z_l
$
with the overall primal solution:
$ 	{x}^*= x_p $.
Here, $x_{l-1}^*$ is the optimal primal of levels $1$ to $l-1$, and $N_{l-1}\in\mathbb{R}^{n\times n_r}$ is a nullspace basis of the vectors $a_k\in\mathbb{R}^n$ ($a_{k}^TN_{l-1}=0$; $N_0=I$). A vector $a_{d/i/j}$ or matrix $A_{d/i/j}$ marked by $\widetilde{A}_{d/i/j}$ represents the projected value $\widetilde{A}_{d/i/j}\coloneqq A_{d/i/j} N_{l-1}$ (or $\widetilde{H}_l\coloneqq N_{l-1}^TH_lN_{l-1}$). The value $\breve{b}_d$ denotes the nullspace projection offset from previous priority levels: $\breve{b}_d \coloneqq b_d - a_d^Tx_{l-1}^*$. ${z}_l\in\mathbb{R}^{n_r}$ is the projected primal of level $l$, with $n_r<n$ for efficiency. 

Below are the primal and dual residuals $K_{\lambda_{\Lambda}}$ and $K_{q_l}$ as the derivative $K_l\coloneqq\nabla\mathcal{L}_l$ of the Lagrangian $\mathcal{L}_l$ of a level $l$ of~\ref{eq:shqp}~\cite{pfeiffer2026}. 
For convenience, we define the placeholder $\Lambda$
\begin{align}
	\Lambda\in\{d^{\pm},d,i,j\}
\end{align}
such that $\lambda_\Lambda$ denotes the Lagrange multiplier
associated with the corresponding constraint class.
Specifically, $\lambda_\Lambda$ represents
$\lambda_{d^\pm}$, $\lambda_d$, $\lambda_i$, or $\lambda_j$
for $\Lambda=d^{\pm},d,i,j$, respectively.
Additionally, the constraint vectors $\ta_d$, $\ta_i$, and $\ta_j$ are stacked row-wise into the matrices \(\tA_d\), \(\tA_i\), and \(\tA_j\), according to their respective index sets \(\mathbb D_l\), \(\mathbb C_l\), and $\mI_{\cup l-1}$.
The $\ell_0$-norm auxiliary constraints $\bD_l^{\pm}$, the inequality constraints in $\bD_l$ and $\bC_l$, and $\mI_{\cup l-1}$ are relaxed by positive slack variables $w_{\Lambda}\geq 0$ (marked by checkered boxes):
\begin{align}
&K_{{\lambda_l}} \hspace{-0pt}=
\BIN  
 ({t}_{d^+}) - ({v}_{d^+})  - (w_{d^+})\\
({t}_{d^-}) + ({v}_{d^-})  - (w_{d^-})\\ 
	-\tA_d{z}_l + (\bb_d) +  ({v}_d) \stripedbox{$\displaystyle{+  (w_d)}$}\\
-\tA_{i}{z}_l + (\bb_{i}) +  ({v}_{i}) \stripedbox{$\displaystyle{+  (w_{i})}$}\\
-\tA_{j}z_l + (\bb_{j}) + (w_{j})
\BOUT\\
	&K_{{q_l}} 	\hspace{-0pt}=	\BIN \widetilde{H}_l{z}_l	\hspace{-0pt} +	\hspace{-0pt} N_{l-1}^T {H}_lx_l    
	\hspace{-0pt}-	 \tA_d^T({\lambda}_d)
	-	 \tA_{i}^T({\lambda}_{i})		\hspace{-0pt} - 	\hspace{-0pt} \tA_{j}^T({\lambda}_{j}) \\
 	({\lambda}_{d^+}) \hspace{-0pt} - \hspace{-0pt} ({\lambda}_{d^-}) \hspace{-0pt}+\hspace{-0pt} ({\lambda}_d)\\
	(\omega_d) \hspace{-0pt}-\hspace{-0pt} ({\lambda}_{d^+}) \hspace{-0pt}-\hspace{-0pt} ({\lambda}_{d^-}) \\
	({v}_{i}) \hspace{-0pt} - \hspace{-0pt} ({\lambda}_{i}) \BOUT
\end{align}

	\section{An ADMM for SHQP}
	\label{sec:nadm}

ADMM solves problems of the following form~\cite{osqp}:

\begin{align}
		\min_{q_l,s_l,\hz_l} \quad& \frac{1}{2}q_l^T{J}_lq_l + g_l^Tq_l
		+ \mathrm{I}_+(s_l)
	\nonumber
	\\
	\text{s.t.}\quad& B_lq_l + C_ls_l = c_l,\quad z_l = \hz_l\label{eq:admmdhlsp}
\end{align}
Applying it to \ref{eq:shqp}, we get the following variable and split variable vectors $q_l$ and $s_l$  (excluding $\hz_l$):
\begin{align}
	q_l &\coloneqq
	\BIN z_l^T & (v_d)^T & (t_d)^T & (v_i)^T \BOUT^T\nonumber\\
	s_l&\coloneqq
	\BIN (w_{d^{\pm}})^T
	& (w_d)^T
	& (w_i)^T
	& (w_j)^T \BOUT^T\nonumber
\end{align}
The slack variables relax inequality constraints, while the operator $\mathrm{I}_+$ projects them onto the set of positive numbers $\mathcal{C}_+$.
The matrices and vectors are composed as:

\scalebox{0.9}{\parbox{1\columnwidth}{
	\begin{align}
	{J}_l&\coloneqq  \BIN \widetilde{H}_l & 0 & 0 & 0\\ 0 & 0 & 0 & 0\\ 0 & 0 & 0 & 0\\ 0 & 0 & 0 & I\BOUT,\quad
	g_l\coloneqq \BIN 0 \\ 0 \\ (\omega_d) \\ 0\BOUT\nonumber
\end{align}}}
\scalebox{0.9}{\parbox{1\columnwidth}{
		\begin{align}
	B_l&= 
	\BIN B_{d^+} \\
	B_{d^-}\\ 
	B_d\\
	B_{i}\\
	B_{j}\\
	\BOUT\coloneqq
	\BIN 
	0 & -I & I & 0\\
	0 & I & I & 0\\
	\tA_d & -I & 0 & 0\\
	\tA_i & 0 & 0 & -I\\
	\tA_j & 0 & 0 & 0\BOUT, 
		c_l = \BIN 
		(c_{d^+}) \\
		(c_{d^-}) \\
		(c_d) \\
		(c_{i}) \\
		(c_{j})
		\BOUT\coloneqq \BIN  0 \\ 0 \\ (\bb_d) \\ 	(\bb_i)\\ 	(\bb_j) \BOUT\nonumber
		\end{align}}}
		\scalebox{0.9}{\parbox{1\columnwidth}{
		\begin{align}
	C_l&=\BIN 
	C_{d^+} \\
	C_{d^-}\\ 
	C_d\\
	C_{i}\\
	C_{j}\\
	\BOUT\coloneqq
	\BIN 
	-I & 0 & 0 & 0 & 0\\
	0 & -I & 0 & 0 & 0\\
	0 & 0 & \stripedbox{$\displaystyle{I}$}   & 0 & 0\\
	0 & 0 & 0 & \stripedbox{$\displaystyle{I}$}   & 0\\
	0 & 0 & 0 & 0 & I \BOUT\nonumber
\end{align}}}

The augmented Lagrangian $\mathcal{L}_{l}^{\rho}$ of level $l$ of~\eqref{eq:admmdhlsp} is:
\begin{align}
	&\mathcal{L}_{l}^{\rho}(q_l,s_l,\hz_l,(\lambda_{\Lambda}))\\
	&= \frac{1}{2}q_l^T{J}_lq_l+g_l^Tq_l + \mathrm{I}_+(s_l)	+ \frac{\sigma}{2}\Vert z_l - \hz_l\Vert^2_2  \nonumber\\
	&+\sum_{\Lambda\in\{d^{\pm},d,i,j\}} \frac{\rho_{\Lambda}}{2} \left\Vert B_{\Lambda}q_l + C_{\Lambda}s_l - (c_{\Lambda}) + \frac{1}{\rho_{\Lambda}}  (\lambda_{\Lambda}) \right\Vert_2^2 \nonumber
\end{align}
The parameters $\sigma\approx 1\cdot 10^{-5}$ and $\rho_{\Lambda}=\rho_l\overline{\rho}_{\Lambda}$  are the individual step-size parameters for each individual constraint $z_l = \hz_l$ and $\Lambda\in\{d^{\pm},d,i, j\}$. The component $\rho_l$ is variable and its update rule is described in Sec.~\ref{sec:rhoupdate}, while $\overline{\rho}_{\Lambda}$ are fixed hand-tuned values for each different constraint set in $\Lambda$ ($\overline{\rho}=100$ for equalities, $\overline{\rho}=1$ for inequalities).

The ADMM computes the primal (see Sec.~\ref{sec:primal}), the split variables (see Sec.~\ref{sec:split}),  and the Lagrange multipliers (see Sec.~\ref{sec:dualascent}) in alternating fashion.
Once the alternating steps of level $l$ have converged with $\Vert K_l\Vert_2<\eta$, the active constraint sets $\mA_{l^*}$ and $\mA_l$ corresponding to activity in $\bD_{l}$, $\bC_{l}$, and $\mI_{\cup l-1}$ are composed.  
The algorithm then continues with the ADMM of the next level $l+1$ projected into the nullspace $N_l$ of the new active set  $\mA_{\cup l} = \mA_{\cup l-1}\cup\mA_l$. The remaining inactive constraints are contained in the updated inactive set $\mI_{\cup l}$. More details on convex hierarchical programming can be found in~\cite{pfeiffer2021}.

\subsection{Primal update}
\label{sec:primal}

The parenthesized superscript $(+)$ (abbreviation for \mbox{$(\iota+1)$}) denotes values of the next ADMM iteration based on the values of the current one $(\iota)$ (omitted for readability). The primal update follows from:
\begin{align}
	&\nabla_{q_l^{(+)}} \mathcal{L}_{l}^{\rho}
	(q_l^{(+)},s_l,\hz_l,(\lambda_{\Lambda}))\nonumber\\
	&=\BIN \widetilde{H}_l+\sigma I \\ &0\\ &&0  \\ &&&I\BOUT q_l^{(+)}+g_l\\
	&+\hspace{-10pt}\sum_{\Lambda\in\{d^{\pm},d,i,j\}}
		\hspace{-15pt}
		B_{\Lambda}^T\rho_{\Lambda}
	\left(B_{\Lambda}q_l^{(+)}+C_{\Lambda}s_l-(c_{\Lambda})
	+\frac{1}{\rho_{\Lambda}}(\lambda_\Lambda)\right)\hspace{-2pt}=\hspace{-2pt}0\nonumber
\end{align}
This can be rewritten into the linear system:
\begin{align}
	K_q^{\rho} q_l^{(+)} = -k_q^{\rho}
\end{align}
where:
\begin{align}
	K_q^{\rho}
	=&
	\BIN 	\widetilde{H}_l + \sigma I_z + M & P^T & 0 & L^T\\
	P & S & 0 & 0\\
	0 & 0 & R & 0\\
	L & 0 & 0 & Y
	\BOUT
	\label{eq:admmLagHess}
\end{align}
and:
\begin{align}
&	M\coloneqq 
\rho_d\tA_d^T\tA_d
+
\rho_{i}\tA_{i}^T\tA_{i}
 + \rho_{j}\tA_{j}^T\tA_{j} \label{eq:KGN}\\\
&P\coloneqq	-\rho_d\tA_d,\quad S\coloneqq \rho_dI + \rho_{d^{\pm}}2I ,\quad R\coloneqq \rho_{d^{\pm}}2I\\
& Y\coloneqq \rho_{i}I,\quad L\coloneqq	-\rho_{i}\tA_i
\end{align}
The matrices $S, R, Y\succ0$ are positive-definite and invertible. The overall matrix $K_q^{\rho}\succ 0$ is then positive-definite as well for $\sigma > 0$ and positive-definite ${H}_l \succ 0$. Note that computing $M$ is the most computationally demanding operation of the algorithm, but only needs to be done once, even in case of $\rho_l$ updates due to the separable nature of $\rho_{\Lambda}=\rho_l\overline{\rho}_{\Lambda}$.

The right hand side $k_{q_l}^{\rho}$ writes as:
\begin{align}
	k_{{z_l}}^{\rho} 	\hspace{-0pt}&=	\textstyle \sum_{\Psi\in \{d,i,j\}} \rho_{\Psi}\tA_{\Psi}^T\left(-(\bb_{\Psi}) + (w_{\Psi}) + \frac{1}{\rho_{\Psi}}(\lambda_{\Psi})\right) \\
	k_{{v}_d}^{\rho} &= 	
		\rho_d\left(\bb_d \stripedbox{$\displaystyle{- w_d}$} -\frac{1}{\rho_d}\lambda_d\right) \\
		&+ \rho_{d^{\pm}}\left(w_{d^+} - w_{d^-}  - \frac{1}{\rho_{d^{\pm}}}\lambda_{d^+} - \frac{1}{\rho_{d^{\pm}}}\lambda_{d^-}\right) \nonumber\\
	  k_{{t}_d}^{\rho} &= 		\omega_d + \rho_{d^{\pm}}\left(-w_{d^+} - w_{d^-} 	+ \frac{1}{\rho_{d^{\pm}}}\lambda_{d^+} +  	 \frac{1}{\rho_{d^{\pm}}}\lambda_{d^-}\right)\\ 
	k_{{v}_{i}}^{\rho} &= 	
	\rho_{i}\left(\bb_{i} \stripedbox{$\displaystyle{- w_{i}}$} -\frac{1}{\rho_{i}}\lambda_{i}\right)
\end{align}
 The vector $(t_d)^{(+)}$ can be directly  computed by:
\begin{align}
	(t_d)^{(+)} &= -R^{-1} (k_{t_d}^{\rho})
\end{align}
Furthermore, the following substitutions of $(v_d)^{(+)}$ and $(v_{i})^{(+)}$ in dependence of ${z}_l^{(+)}$ can be computed by:
\begin{align}
	(v_d)^{(+)} &= S^{-1}\left(-P{z}_l^{(+)}  -(k_{{v}_d}^{\rho})\right)\\
	(v_{i})^{(+)} &= Y^{-1}\left(-L{z}_l^{(+)}  -(k_{{v}_{i}}^{\rho})\right)
\end{align}
Finally, ${z}_l^{(+)}$ is efficiently computed from:
\begin{align}
	K_z^{\rho} z_l^{(+)}  =  \left( P S^{-1} (k_{v_d}^{\rho}) + L^TY^{-1}(k_{v_i}^{\rho}) -k_{z_l}^{\rho}\right)
	\label{eq:solve}
\end{align}
for example using a Cholesky decomposition of the normal-matrix $K_{z_l}^{\rho}\in \mathbb{R}^{n_r\times n_r}$:
\begin{align}
	K_{z_l}^{\rho}\coloneqq\left(\widetilde{H}_l + \sigma I_z + M - P^T S^{-1}P - L^T Y^{-1}L\right) 
	\label{eq:normalmatrix}
\end{align}

\subsection{Split variable update}
\label{sec:split}

The split variable update results from the condition ($\Lambda\in\{d^{\pm},d,i,j\}$):
\begin{align}
	&\nabla_{s_l^{(+)},\hz_l^{(+)}} \mathcal{L}_{l}^{\rho}(q_l^{(+)},s_l^{(+)},\hz_l^{(+)},(\lambda_{d^{\pm}}),...)\\ 
	=& \BIN  \rho_{\Lambda}C_{\Lambda}^TC_{\Lambda}s_l^{(+)} +  C_{\Lambda}^T\rho_{\Lambda}(B_{\Lambda}q_l^{(+)} - (c_{\Lambda}) + \frac{1}{\rho_{\Lambda}}(\lambda_{\Lambda}))\\
	\hz_l^{(+)} - z_l^{(+)}\BOUT = 0\nonumber
\end{align}
This leads to the updates $\hz_l^{(+)} = z_l^{(+)}$ and:
\begin{align}
	w_{d^+}^{(+)} &=  v_{d^+}^{(+)} - {t}_{d^+}^{(+)} - \frac{1}{\rho_{d^{\pm}}}\lambda_{{d^+}}\\
w_{d^-}^{(+)} &=  -v_{d^-}^{(+)} - {t}_{d^-}^{(+)} - \frac{1}{\rho_{d^{\pm}}}\lambda_{d^-}\\
w_{d/i}^{(+)} &=	-\tA_{d/i} z_l^{(+)} + v_{d/i}^{(+)} + \bb_{d/i} - \frac{1}{\rho_{d/i}}\lambda_{d/i}\\
w_{j}^{(+)} &=	-\tA_{j} z_l^{(+)} + \bb_{j} - \frac{1}{\rho_{j}}\lambda_{j}
\end{align}
Finally, the operator $\mathrm{I}_+$ is realized by:
\begin{align}
	w_{\Lambda}^{(+)} \leftarrow \max(0, w_{\Lambda}^{(+)})
\end{align}

\subsection{Dual update}
\label{sec:dualascent}

The Lagrange multipliers are updated by dual ascent~\cite{boyd2011}, which writes as ($\Lambda\in\{{d^{\pm}},d,i,j\}$):
\begin{align}
	(\lambda_{\Lambda})^{(+)} = (\lambda_{\Lambda}) + \rho_{\Lambda}(B_{\Lambda}q_l^{(+)} + C_{\Lambda}s_l^{(+)} - c_{\Lambda})
\end{align}
Specifically, this leads to the updates:
\begin{align}
\lambda_{d^+}^{(+)} &= \lambda_{d^+} + \rho_{d^{\pm}}\left(-v_{d^+}^{(+)} + {t}_{d^+}^{(+)} + w_{d^+}^{(+)}\right)
\\
\lambda_{d^-}^{(+)} &= \lambda_{d^-} + \rho_{d^{\pm}}\left(v_{d^-}^{(+)} + {t}_{d^-}^{(+)} + w_{d^-}^{(+)}\right)\\
	\lambda_{d/i}^{(+)} &= \lambda_{d/i} +  \rho_{d/i}\left(\tA_{d/i}z_l^{(+)}  - v_{d/i}^{(+)} \stripedbox{$\displaystyle{+ w_{d/i}^{(+)}}$}  - \bb_{d/i}\right)\\
	\lambda_{j}^{(+)} &= \lambda_{j} +  \rho_{j}\left(\tA_{j}z_l^{(+)}  + w_{j}^{(+)} - \bb_{j}\right)
\end{align}

\subsection{Computational comparison}

\begin{table*}[t]
	\centering
	
	\resizebox{\textwidth}{!}{%
		\begin{tabular}{lccccc}
			\hline
			\textbf{Operation}
			&
			\textbf{$\mathcal{N}$\hspace{-2pt}ADM}
			&
			\textbf{$\mathcal{N}$IPM}~\cite{pfeiffer2026}
			&
			\textbf{PIQP}$^*$~\cite{piqp}
			&
			\textbf{OSQP}$^*$~\cite{osqp}
			&
			\textbf{qpOASES}~\cite{Ferreau2014}
			\\
			\hline
			
			Primal variables
			&
			$n$
			&
			$n$
			&
			$n+2m$
			&
			$n+2m$
			&
			$n+2m$
			\\
			
			KKT matrix dim.~\eqref{eq:normalmatrix}
			&
			$n^2$
			&
			$n^2$
			&
			$(n+2m)^2$
			&
			$(n+5m)^2$
			&
			$a\times(n+2m)$
			\\
			
			Matrix-vector ops.
			&
			$\mathcal{O}(\iota mn)$
			&
			$\mathcal{O}(\iota mn)$
			&
			$\mathcal{O}(\iota mn)$
			&
			$\mathcal{O}(\iota mn)$
			&
			$\mathcal{O}\!\left(
			\iota 3m(n+2m)
			\right)$
			\\
			
			Normal-matrix eval.~\eqref{eq:normalmatrix}
			&
			$\mathcal{O}(mn^2)$
			&
			$\mathcal{O}(\iota mn^2)$
			&
			$\mathcal{O}(mn^2)$
			&
			--
			&
			--
			\\
			
			Matrix factorization~\eqref{eq:solve}
			&
			$\mathcal{O}(n^3)$
			&
			$\mathcal{O}(\iota n^3)$
			&
			$\mathcal{O}\!\left(
			\iota(n+2m)^3
			\right)$
			&
			$\mathcal{O}\!\left(
			(n+5m)^3
			\right)$
			&
			$\mathcal{O}\!\left(a^2
			(n+2m)+\iota a
			(n+2m)
			\right)$
			\\
			
			Back substitution~\eqref{eq:solve}
			&
			$\mathcal{O}(\iota n^2)$
			&
			$\mathcal{O}(\iota n^2)$
			&
			$\mathcal{O}\!\left(
			\iota(n+2m)^2
			\right)$
			&
			$\mathcal{O}\!\left(
			\iota(n+5m)^2
			\right)$
			&
			$\mathcal{O}\!\left(
			\iota a(n+2m)
			\right)$
			\\
			\hline
			
			\textbf{Total}
			&
			$\mathbf{\mathcal{O}}\!\left(
			\mathbf{mn^2+n^3+}
			\iota\mathbf{n(m+n)}
			\right)$
			&
			$\mathbf{\mathcal{O}}\!\left(
			\iota\left(\mathbf{mn^2+n^3}\right)
			\right)$
			&
			$\mathbf{\mathcal{O}}\!\left(
			\iota\mathbf{(n+2m)^3}
			\right)$
			&
			$\mathbf{\mathcal{O}}\!\left(
			\mathbf{(n+5m)^3}
			+\iota\mathbf{(n+5m)^2}
			\right)$
			&
			$\mathbf{\mathcal{O}}\!\left(
			\mathbf{(n+2m)}
				\left(
				\mathbf{a^2} + \iota	\mathbf{(m+a)}
				\right)
			\right)$
			\\
			
			\hline
		\end{tabular}%
	}
		\caption{Computational cost over $\iota$ iterations for
		$m\coloneqq\vert\bD_{1}\vert$ L0-equality constraints
		($A_d\in\mathbb{R}^{m\times n}$).
		Solvers marked with $^*$ are implemented using sparse linear algebra,
		such that the computational cost is typically governed by sparsity
		rather than the raw matrix dimensions.
		For qpOASES, $\iota$ denotes the number of active-set changes, and $a \leq 3m$ is the number of linearly independent active constraints.}
	\label{tab:complexity}
\end{table*}
Each iteration of our solver (referred to as \mbox{$\mathcal{N}$\hspace{-2pt}ADM}), $\mathcal{N}$IPM~\cite{pfeiffer2026}, PIQP~\cite{piqp}, and OSQP~\cite{osqp} involves the sets of operations summarized in Tab.~\ref{tab:complexity}. Due to the lack of elimination of $\ell_0$-norm auxiliaries, the complexities of both PIQP and OSQP are strongly influenced by the number of constraints $m\gg1$ (short of cubic due to sparse linear algebra implementations), which is typically large in decision-making. On the other hand, both $\mathcal{N}$\hspace{-2pt}ADM and $\mathcal{N}$IPM are linearly dependent on $m$, while \mbox{$\mathcal{N}$\hspace{-2pt}ADM} needs to conduct the expensive normal-matrix assembly~\eqref{eq:normalmatrix} of order $\mathcal{O}(mn^2)$ only once, even if there are step-size parameter $\rho_l$ updates.

\section{Further ADMM components}
\label{sec:admmcomponents}

\subsection{Equilibration}
\label{sec:precond}

We apply Ruiz equilibration~\cite{ruizscaling} for preconditioning in order to reduce the number of iterations $\iota$. Here, matrix rows and columns are iteratively scaled by their inverse square root maximum absolute value until they equal to 1. For QPs, it is suggested to scale the  Lagrangian Hessian $\mathcal{H}_l\coloneqq\nabla_{q_l,\lambda_{l}}K_l$~\cite{osqp}. In the specific case of~\ref{eq:shqp}s, this leads to the scaled  matrix $\overline{\mathcal{H}}_l$ (symmetry indicated by $\star=(\cdot)^T$; $\Sigma$ is a diagonal matrix containing the scaling factors):
\begin{align}
	\overline{\mathcal{H}}_l
	&= \Sigma \mathcal{H}_l \Sigma=
	\label{eq:Hbar} \\[-0.2em]
	&
	\scalebox{0.77}{$
		\left[
		\begin{array}{lccccccccc}
			& {\color{gray} \overline{z}_l}
			& {\color{gray} (\overline{v}_d)}
			& {\color{gray} (\overline{t}_d)}
			& {\color{gray} (\overline{v}_i)}
			& {\color{gray} (\overline{\lambda}_{d^+})}
			& {\color{gray} (\overline{\lambda}_{d^-})}
			& {\color{gray} (\overline{\lambda}_d)}
			& {\color{gray} (\overline{\lambda}_i)}
			& {\color{gray} (\overline{\lambda}_j)}
			\\
			{\color{gray} \overline{z}_l}
			& \overline{\widetilde{H}}_l
			& & & & & & \star&\star & \star
			\\
			{\color{gray} (\overline{v}_d)}
			& & & & &\star & \star & \star & 
			\\
			{\color{gray} (\overline{t}_d)}
			& & & & &\star & \star
			\\
			{\color{gray} (\overline{v}_i)}
			& & & & D & & & & \star  &
			\\
			{\color{gray} (\overline{\lambda}_{d^+})}
			& & -V^+ & T^+ & & & & & &
			\\
			{\color{gray} (\overline{\lambda}_{d^-})}
			& & V^- & T^- & & & & & &
			\\
			{\color{gray} (\overline{\lambda}_d)}
			& \overline{\tA}_d & -E & & & & & & &
			\\
			{\color{gray} (\overline{\lambda}_i)}
			& \overline{\tA}_i & & & -F & & & & &
			\\
			{\color{gray} (\overline{\lambda}_j)}
			& \overline{\tA}_j & & & & & & & &
		\end{array}
		\right]
		$}
	\notag
\end{align}
Gray entries indicate the individual Hessian rows and columns as derivatives with respect to the given vector.
Overlined matrices and vectors are the equilibrated equivalents of their original counterparts. The matrices $E$, $F$ $D$, $V^{\pm}$, and $T^{\pm}$ are scaled diagonal matrices.

In the following, we make an observation which leads to simplifications in the scaled version of $\overline{K}_q^{\rho}$~\eqref{eq:admmLagHess}, as off-diagonal elements of the form $-\rho_{d^{\pm}}T^+V^+ + \rho_{d^{\pm}}T^-V^-$ are canceled out.
We use the operator $\Gamma$ which returns a vector containing the maximum value of each row of a matrix, or of a row / column block $(\cdot)$ of $\overline{\mathcal{H}}_l$.
\begin{prop}
	Throughout the Ruiz equilibration of $\nabla^2_{q,\lambda}\mathcal{L}$, it holds that $E\leq 1$ and therefore $V^{\pm}=T^{\pm}=I$.
\end{prop}
\begin{proof}
	Initially, $E=V^{\pm}=T^{\pm}=I$, so $\Gamma((\overline{v}_d))=\Gamma((\overline{\lambda}_{d^{\pm}}))=1$ follow immediately. At the same time, $\Gamma((\overline{\lambda}_d))\leq 1$ since for any maximum row value $\Gamma(\overline{\tA}_d)<1$, we have that $\Gamma(E)=1 > \Gamma(\overline{\tA}_d)$. We then clearly have $E\leftarrow\diag(\sqrt{(\Gamma(\overline{\lambda}_d))})^{-1}E\diag(\sqrt{(\Gamma(\overline{v}_d))})^{-1} \leq 1$. 
	In the next Ruiz iteration, $\Gamma((\overline{v}_d))=\Gamma((\overline{\lambda}_{d^{\pm}}))=1$ still holds due to the unchanged $V^{\pm}=T^{\pm}=I$ and $E\leq 1$. By induction over the Ruiz iterations, the claims follow. 
\end{proof}

\subsection{Step-size parameter $\rho_l$ updates}

\label{sec:rhoupdate}

We update the step-size parameter $\rho_l$ (initially, $\rho_l=0.1$) by the update rule presented in~\cite{osqp}, where the operator $\sum \Vert \cdot\Vert_{\infty}$ returns the absolute maximum value of all additive components in $K_{\lambda_l}$ and $K_{q_l}$:
\begin{align}
	\rho_l \leftarrow \rho_l 
	\sqrt{
		\frac{
			\Vert K_{\lambda_l}\Vert_{\infty} / \max\left(\sum \Vert K_{\lambda_l}\Vert_{\infty}\right)}{
			\Vert K_{q_l}\Vert_{\infty} / \max\left(\sum \Vert K_{q_l}\Vert_{\infty}\right)
	}}\nonumber
\end{align}
It balances reduction in the normalized primal and dual residuals $K_{\lambda_l}$ and $K_{q_l}$, respectively. 
If $\rho_l$ is updated, 	$K_{z_l}^{\rho}$	\eqref{eq:normalmatrix} needs to be updated and re-factorized.

\subsection{Over-relaxation}
In over-relaxed ADMM, the relaxed quantity
$B_lq_l^{(+)} \leftarrow \alpha B_lq_l^{(+)}-(1-\alpha)(C_ls_l-c_l)$
is used in the split-variable and dual-variable updates
(Sec.~\ref{sec:split} and~\ref{sec:dualascent}) in order to reduce the number of iterations $\iota$~\cite{Ghadimi2014}. The parameter
$\alpha\in(0,2)$ is the over-relaxation parameter (typically $\alpha=1.6$~\cite{osqp}).

\subsection{Warm-starting}
Between problem instances, for each priority level $l$ the previously found step-size parameter $\rho_l^-$, and the unscaled primal, split, and dual variables $q_l^-$, $s_l^-$, and $\lambda_{\Lambda}^-$, respectively, are stored. A level $l$ is only warm-started if the current active and inactive constraint sets $\mA_{\cup l-1}$ and $\mI_{\cup l-1}$ are identical with the ones of the previous problem instance at that level $\mA_{\cup l-1}^-$ and $\mI_{\cup l-1}^-$, while standard initialization is used otherwise.

\section{Separable decision functions}
\label{sec:sepdm}

The separable structure of decision functions $f_d(\chi)$ of the form as in~\eqref{eq:dmfct} can be exploited for computational gain.
One row of the Jacobian $J_d$ of $f_d(\chi)$ writes as:
\begin{align}
	a_d^T
	&=  \hat{g}_d^TG\quad \text{with} \quad \hat{g}_d\coloneqq g(\chi) - g_d,\quad G \coloneqq \frac{\partial g(\chi)}{\partial\chi}
\end{align}
We use $\hat{G}\coloneqq \BIN\cdots & \hat{g}_d &\cdots \BOUT\in\mathbb{R}^{u\times\vert\mathbb{D}_l\vert}$.
Then, a matrix product evaluated in~\eqref{eq:KGN} can be rewritten to:
	\begin{align}
		A_d^TA_d
		&= G^T\hat{G}\hat{G}^TG
	\end{align}
Evaluating this matrix naively by $\widetilde{G}^T\widetilde{G}$ with $\widetilde{G}=\hat{G}^TG$ is of order $O(\vert\mathbb{D}_l\vert un + \vert\mathbb{D}_l\vert n^2)$, so the large factor $\vert\mathbb{D}_l\vert\gg 1$ is amplified by $n^2$.
Instead, by first computing $\hat{G}\hat{G}^T$, the overall evaluation is of order $O(\vert\mathbb{D}_l\vert u^2 + un^2)$, where $\vert\mathbb{D}_l\vert$ is amplified by only $u^2$. The factor $u$ is typically small in robotics, for example  $u=3$ for translational reaching.

\section{Evaluation}
\label{sec:eval}

We compare the proposed solver $\mathcal{N}$\hspace{-2pt}ADM (cold and warm) with $\mathcal{N}$IPM~\cite{pfeiffer2026} and PIQP~\cite{piqp} (both no warm-start), and OSQP~\cite{osqp} and qpOASES~\cite{Ferreau2014} (both cold and warm). PIQP, OSQP, and qpOASES solve \ref{eq:shqp} directly without eliminating the auxiliary variables $(t_d)$ and the related constraints. Except for the walking scenario (Sec.~\ref{sec:g1walk}), all solvers terminate at $\Vert K_l\Vert_{\infty} < 1\cdot10^{-3}$. Timings purely reflect the solve times, while problem setup / update times, ... are excluded. 
The robot \ref{eq:shqp}s are composed by Pinocchio~\cite{carpentier2019pinocchio}, with PD controllers expressed in the velocity domain~\cite{pfeiffer2023} (Sec.~\ref{sec:drone} and Sec.~\ref{sec:g1walk}). Evaluations are run on an Intel Core Ultra 9 185H CPU with 64~GB RAM.

\subsection{Large-scale~LPs}

\begin{figure}[htp!]
	\includegraphics[width=1 \columnwidth]{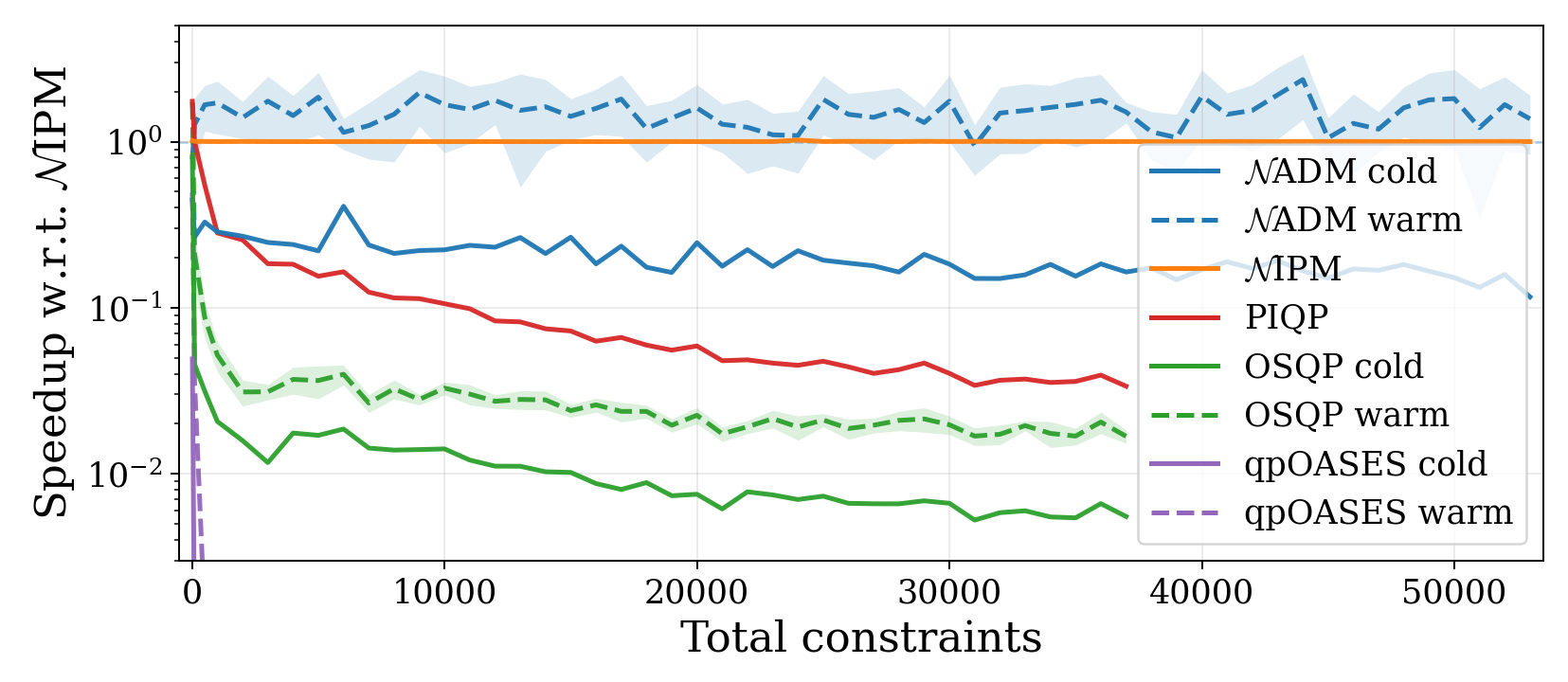}
	\centering
	\caption{Speedup of different solvers with reference to $\mathcal{N}$IPM on large-scale~LPs with $p=1$, $n=30$, and $\vert\bD_1\vert= 10,100,500,1\cdot 10^3,...,5.3\cdot 10^5$ equality constraints.}
	\label{fig:largescaleshqp}
\end{figure}

We first isolate the benefit of exploiting sparse structure over general-purpose solvers based on sparse linear algebra~\cite{piqp,osqp}, independently of hierarchical effects.
For this, we solve large-scale linear programs (LPs) corresponding to the case of $p=1$ and $H_1=0$ in~\ref{eq:shqp}.
Equality constraints are composed as randomized, dense matrices with $n=30$ and $A_d\in\mathbb{R}^{\vert\bD_1\vert\times n}$ ($\vert\bD_1\vert = 10,100,500,1\cdot 10^3,...,5.2\cdot 10^5$, increment: $1000$). For each $\vert\bD_1\vert$, we repeatedly ($10\times$) solve slightly disturbed problems  $\tA_d + B$ in order to simulate warm-starting, where $B$ is a randomized dense matrix with elements in the range of $1\cdot10^{-3}\Vert \tA_d\Vert_{\infty}$.

Figure~\ref{fig:largescaleshqp} shows that the proposed warm-started solver \mbox{$\mathcal{N}$\hspace{-2pt}ADM}  achieves a mean speedup of around $21.42 \pm 14.91$\% with respect to $\mathcal{N}$IPM ($123 \pm 29$ and $16\pm1$ iterations, $0.07 \pm 0.05$s and $0.09\pm 0.06$s solve time over all runs for $\mathcal{N}\hspace{-2pt}$ADM and $\mathcal{N}$IPM, respectively). On the contrary, the reference solvers PIQP and OSQP are between one to two orders of magnitude slower. qpOASES is significantly slower due to its dense linear algebra implementation, but exhibits a significant drop in active-set changes  (typically 0 to 3 iterations) after the initial cold solve. While PIQP does not possess explicit warm-starting functionality, its iteration count is low at $3 \pm 0.4$ iterations, which emphasizes the need to exploit the structure of sparse programs. PIQP, OSQP, and qpOASES could only be evaluated up to $3.6\cdot 10^5$ constraints due to their increased memory requirements ($2\times$ variables and $3\times$ more constraints) as no elimination of the sparse auxiliaries takes place.

\subsection{Humanoid inverse kinematic control}

\begin{table}[htp!]
	\centering
	\resizebox{0.8\columnwidth}{!}	{%
		\begin{tabular}{@{} cccccc @{}}  
			\toprule
$\underline{l}$ & & $\ell$ & $f_{d/i}(\chi)\leqq \nu_{d/i},\quad d/i\in\bD_l/\bC_l$ \\
\midrule
$\underline{1}$ & Vel. lim. / trust region& 2 & {$ \vert \dot{\mathcal{Q}}\vert \leq 0.01$}   \\
\rowcolor{gray!20}$\underline{2}$ & Angle lim.& 2 & {$  \underline{{\mathcal{Q}}} \leq {\mathcal{Q}} \leq \overline{{\mathcal{Q}}} $}   \\
$\underline{3}$ & Stance & 2 & {$f_{\text{LF/RF/LH}}= \nu_{\text{LF/RF/LH}}$}   \\
\rowcolor{gray!20}$\underline{4}$ & Coll. av. & 2 & $f_{\text{coll}}(\mathcal{Q})\geq \nu_{\text{coll}}$   \\
$\underline{5}$& Goal DM  & 0 & $\Vert f_{\text{RH}} - f_{\text{RH},d}\Vert_2^2=\nu_{\text{RH},d}$   \\
			\bottomrule
		\end{tabular}
	}
	\caption{Inverse kinematic control of humanoid robot, task hierarchy.}
	\label{tab:ikvel}
\end{table}
\begin{figure}[htp!]
	\includegraphics[width=1 \columnwidth]{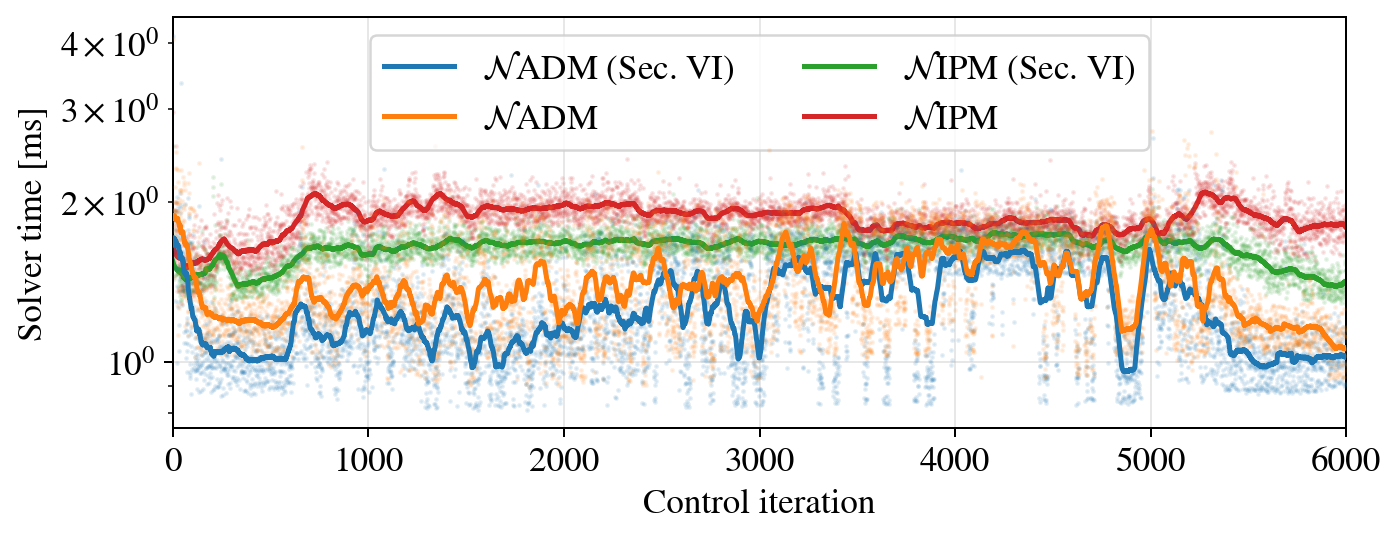}
	\centering
	\caption{Inverse kinematic control of humanoid robot, solve times per control 	iteration.}
	\label{fig:shikcg1_time}
	\end{figure}
	
	Here, we evaluate warm-starting and the linear algebra implementation for separable decision functions as outlined in Sec.~\ref{sec:sepdm}. A simulation is designed where the velocity controlled humanoid robot Unitree G1 $\chi=\mathbb{R}^{29}$ needs to touch as many as possible passing objects out of 500.

Figure~\ref{fig:shikcg1_time} shows that after 200 iterations of warm-starting $\mathcal{N}$\hspace{-2pt}ADM (maximum solve time: 4.1~ms, $\mathcal{N}$IPM: 3.0~ms), the solver reaches a steady level of around $1.3 \pm 0.3$~ms ($\mathcal{N}$IPM: $1.6 \pm 0.1$), with $\mathcal{N}$\hspace{-2pt}ADM being 23.5\% faster than $\mathcal{N}$IPM over all control iterations. $\mathcal{N}$\hspace{-2pt}ADM (Sec.~\ref{sec:sepdm}) and $\mathcal{N}$IPM (Sec.~\ref{sec:sepdm}), which exploit the structure of the decision functions according to~Sec.~\ref{sec:sepdm}, improve the solve time per control iteration by 10\% and 12\% over their baselines, respectively.

\subsection{Drone control}
\label{sec:drone}
\begin{table}[htp!]
	\centering
	\resizebox{0.8\columnwidth}{!}	{%
		\begin{tabular}{@{} cccccc @{}}  
			\toprule
			$\underline{l}$ & & $\ell$ & $f_{d/i}(\chi)\leqq \nu_{d/i},\quad d/i\in\bD_l/\bC_l$ \\
			\midrule
			$\underline{1}$ & Vel. lim. / trust region& 2 & {$ \vert \dot{\mathcal{Q}}\vert \leq 0.01$}   \\
			\rowcolor{gray!20}$\underline{2}$& Force lim. & 2 & {$\underline{\mathcal{F}} \leq \mathcal{F} \leq \overline{\mathcal{F}}$} \\ 
		$\underline{3}$ & Coll. av. & 2 & $f_{\text{coll}}(\mathcal{Q})\geq \nu_{\text{coll}}$   \\
		\rowcolor{gray!20}$\underline{4}$& Dynamics & 2 & $f_{\text{dyn}}(\mathcal{Q})=\nu_{\text{dyn}}$\\
			$\underline{5}$& Goal DM  & 0 & $\Vert \mathcal{P} - P_d\Vert_2^2=\nu_d$   \\
			\rowcolor{gray!20}$\underline{6}$& Goal tracking  & 2 & $f_{\text{Drone}}(\mathcal{Q})-\mathcal{P}=\nu_i$   \\
			\bottomrule
		\end{tabular}
	}
	\caption{Drone control, task hierarchy.}
	\label{tab:xarm6shikp}
\end{table}

\begin{figure}[htp!]
	\includegraphics[width=1 \columnwidth]{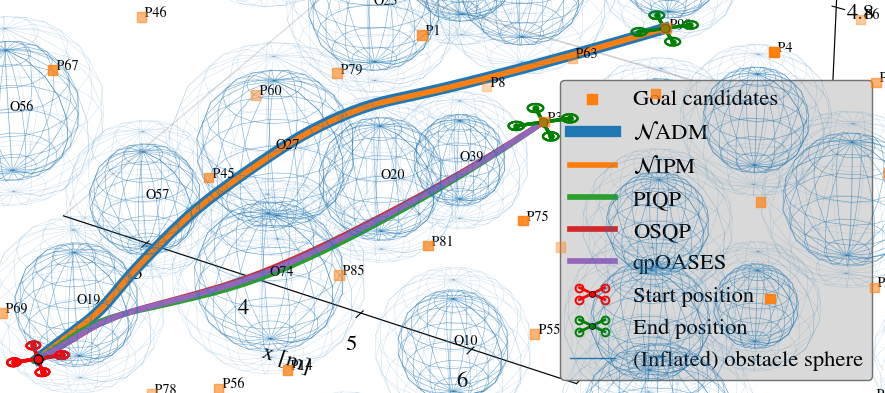}
	\centering
	\caption{Instantaneous hierarchical control of a drone to navigate through an obstacle course, while deciding on one of 100 possible goal locations.}
	\label{fig:dronepath}
\end{figure}

\begin{figure}[htp!]
	\includegraphics[width=1 \columnwidth]{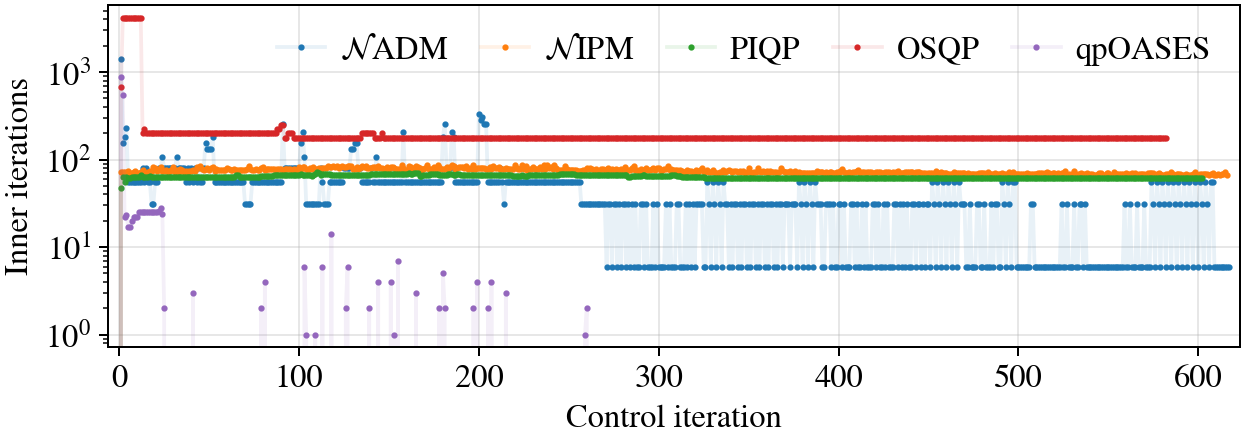}
	\centering
	\caption{Drone control, number of inner iterations per control iteration.}
	\label{fig:droneiter}
\end{figure}

\begin{table}[t]
	\centering
	\begin{tabular}{lccc}
		\hline
		Solver & Inner iter. & Time & 95\textsuperscript{th} perc. \\
		\hline
		$\mathcal{N}$\hspace{-2pt}ADM & $44.3 \pm 69.7$   & $0.24 \pm 0.18$ & $0.33$ \\
		$\mathcal{N}$IPM~\cite{pfeiffer2026} & $74.7 \pm 6.0$    & $0.67 \pm 0.10$ & $0.75$ \\
		PIQP~\cite{piqp}             & $64.0 \pm 3.6$    & $0.86 \pm 0.09$ & $0.97$ \\
		OSQP~\cite{osqp}              & $254.8 \pm 541.5$ & $0.38 \pm 0.52$ & $0.34$ \\
		qpOASES~\cite{Ferreau2014}              & $3.5 \pm 42.9$ & $9.39 \pm 19.32$ & $9.16$ \\
		\hline
	\end{tabular}
		\caption{Drone control, comparison of number of inner iterations and solve time (ms) per control iteration.}
	\label{tab:dronecomp}
\end{table}

In this example, we create an SH-NLP with $p=6$ and $n=9$ for navigation of a drone with linear dynamics $f_{dyn}$ through an obstacle course to choose one of 100 targets (see Fig.~\ref{fig:dronepath}).  The problem is solved with an instantaneous controller with $\Delta t=20$~ms.
The drone is described by $\chi = [\mathcal{P},\mathcal{Q},\mathcal{F}]$, where $\mathcal{P}\in\mathbb{R}^3$ is the current goal location, $\mathcal{Q}\in\mathbb{R}^3$ is the current drone translation~\cite{pfeiffer2026}, and $\mathcal{F}\in\mathbb{R}^3$ is the control force. Since the drone is controlled via a proportional-derivative (PD) controller expressed in the velocity domain~\cite{pfeiffer2023}, imposing the decision-making directly on acceleration commands is less effective and does not accurately reflect the true decision-making error to the physical target. Instead, the decision-making is imposed on a separate goal position $\mathcal{P}$ which chooses one of the goal locations $P_d$. The goal location is then realized by a PD controller on the last priority level. 
Further constraints on velocities (max $0.5$ m/s), control forces (6 N and 18 N max. horizontal and vertical), and collision avoidance complete the hierarchy. 

Figure~\ref{fig:droneiter} and Tab.~\ref{tab:dronecomp} show how \mbox{$\mathcal{N}$\hspace{-2pt}ADM} is warm-started very effectively, where in many iterations the controller applies the same control $\mathcal{F}$ in subsequent iterations. In contrast, warm-started OSQP still requires around 200 iterations, while qpOASES reaches steady-state after around 24 control iterations. 

\subsection{Humanoid whole-body control for walking over stepping stones}
\label{sec:g1walk}

\begin{table}[htp!]
	\centering
	\resizebox{0.8\columnwidth}{!}	{%
			\begin{tabular}{@{} cccccc @{}}  
				\toprule
			$\underline{l}$ & & $\ell$ & $f_{d/i}(\chi)\leqq \nu_{d/i},\quad d/i\in\bD_l/\bC_l$ \\
				\midrule
				$\underline{1}$ & Angle lim. / Vel. lim.& 2 & {$\underline{\mathcal{Q}} \leq \mathcal{Q} \leq \overline{\mathcal{Q}}$, $ \vert \dot{\mathcal{P}}\vert \leq 0.01$ / $ \vert \dot{\mathcal{Q}}\vert \leq 0.01$}   \\
				\rowcolor{gray!20}$\underline{2}$& Torque lim. & 2 & {$\underline{\mathcal{T}} \leq \mathcal{T} \leq \overline{\mathcal{T}}$} \\ 
				$\underline{3}$& Dynamics & 2 & $f_{\text{dyn}}(\mathcal{Q})=\nu_{\text{dyn}}$\\
				\rowcolor{gray!20}$\underline{4}$& Wrench cone  & 2 & {$f_{\text{wc}}(\gamma) \geq \nu_{\text{wc}}$} \\ 
				$\underline{5}$& Stance-foot  & 2 & $f_{\text{LF/RF}}(\mathcal{Q})=\nu_{\text{LF/RF}}$   \\
				\rowcolor{gray!20}$\underline{6}$ & Coll. av. & 2 & $f_{\text{coll}}(\mathcal{Q})\geq \nu_{\text{coll}}$   \\
				$\underline{7}$& DCM / upright torso & 2 & $f_{\text{stab}}(\mathcal{Q},\dot{\mathcal{Q}})\geq\nu_{\text{stab}}$\\
			\rowcolor{gray!20}$\underline{8}$& Goal DM  & 0 & $\Vert \mathcal{P} - P_d\Vert_2^2=\nu_d$   \\
			$\underline{9}$& Swing-foot  & 2 & $f_{\text{LF/RF}}(\mathcal{P},\mathcal{Q})=\nu_{\text{LF/RF}}$   \\
			\rowcolor{gray!20}	$\underline{10}$& Regularization & 2 & $\chi=\nu_i$   \\
				\bottomrule
			\end{tabular}
		}
	\caption{Humanoid whole-body control for walking over stepping stones, task hierarchy.}
	\label{tab:humanoidwalking}
\end{table}

In this experiment, we demonstrate the integration of hierarchical decision-making within a dynamic humanoid walking controller see Fig.~\ref{fig:g1walk}. Both the left and right feet have to make a decision on a candidate set of $\vert \bD_8\vert = 200$ stepping stones to step on next, while the other foot is fixed on the previously chosen stepping stone. Whole-body dynamics, joint torque limits, 6D contact wrench conditions, collision avoidance, and DCM stability constraints~\cite{Englsberger2015} complete the hierarchy given in Tab.~\ref{tab:humanoidwalking}. 
The variable vector $\chi\in\mathbb{R}^{67}$ is composed of the goal location $\mathcal{P}\in\mathbb{R}^3$, joint velocities $\dot{\mathcal{Q}}\in\mathbb{R}^{29}$, joint torques $\mathcal{T}\in\mathbb{R}^{23}$, and left and right foot contact wrenches $\gamma\in\mathbb{R}^{12}$. 

The problem can only be faithfully solved under high-accuracy settings ($\Vert K_l\Vert_{\infty} < 1\cdot10^{-12}$), which is less suitable for first-order methods like \mbox{$\mathcal{N}$\hspace{-2pt}ADM} and OSQP. $\mathcal{N}$\hspace{-2pt}ADM's maximum iteration count is set to $1\cdot 10^{5}$, making its computation time of $880.7 \pm 413.5$~ms per control iteration not competitive with $\mathcal{N}$IPM's and PIQP's runtime of $19.8 \pm 2.7$~ms and $66.9 \pm 8.1$~ms, respectively. This highlights the existence of problem classes which require high-accuracy solution regimes.

\section{Conclusion}

In this work, we have presented an ADMM solver for hierarchical decision-making control, which effectively exploits temporal similarities between nonlinear solver subproblems. Evaluation shows that in low- to moderate-accuracy regimes, the solver achieves speedups of up to 23.5\% over the structure-exploiting IPM~\cite{pfeiffer2026}. At the same time, complex scenarios like humanoid walking expose the limitations of ADMM based solvers when high-accuracy solutions are required.

\section{Acknowledgments}

This work was supported by the RDF-24-02-060 XJTLU Research Development Fund.
		
\balance
		
\bibliographystyle{IEEEtran}
\bibliography{bib}

\end{document}